# FATE: Fast and Accurate Timing Error Prediction Framework for Low Power DNN Accelerator Design

Jeff (Jun) Zhang     Siddharth Garg

New York University

{jz2163, sg175}@nyu.edu

## ABSTRACT

Deep neural networks (DNN) are increasingly being accelerated on application-specific hardware such as the Google TPU designed especially for deep learning. Timing speculation is a promising approach to further increase the energy efficiency of DNN accelerators. Architectural exploration for timing speculation requires detailed gate-level timing simulations that can be time-consuming for large DNNs that execute millions of multiply-and-accumulate (MAC) operations. In this paper we propose FATE, a new methodology for fast and accurate timing simulations of DNN accelerators like the Google TPU. FATE proposes two novel ideas: (i) DelayNet, a DNN based timing model for MAC units; and (ii) a statistical sampling methodology that reduces the number of MAC operations for which timing simulations are performed. We show that FATE results in between $8\times-58\times$ speed-up in timing simulations, while introducing less than 2% error in classification accuracy estimates. We demonstrate the use of FATE by comparing to conventional DNN accelerator that uses 2's complement (2C) arithmetic with an alternative implementation that uses signed magnitude representations (SMR). We show that that the SMR implementation provides 18% more energy savings for the same classification accuracy than 2C, a result that might be of independent interest.

## 1. INTRODUCTION

The recent success of deep learning in a range of machine learning applications [16, 18, 23, 30] has motivated interest in the design of special purpose hardware accelerators for both training and inference of deep neural networks (DNN). Orders of magnitude energy savings compared to CPU or GPU based solutions have been realized by recent silicon prototypes that seek to accelerate the computationally expensive matrix multiplication and convolution operations needed for DNN inference and training using highly parallel arrays of multiply-and-accumulate (MAC) units [4, 5, 8, 15, 20, 22]. Techniques such as zero-skipping and neural network compression seek to further reduce the energy consumption of deep learning accelerators by exploiting special properties of DNNs [1, 5, 17, 21].

In this paper, we are interested in using timing speculation, which has recently been shown as another promising approach to increase the energy efficiency of DNN accelerators [27, 29]. Conventional digital design methodologies pessimistically include significant timing margins (or guardbands) to guarantee correct functionality in the presence of process, voltage, and temperature (PVT) variations. The idea behind timing speculation is to optimistically execute a chip at a lower voltage thus reduce energy at the expense of higher delay and consequently occasional timing errors [9]. Timing errors can either be allowed to propagate if the application is itself error tolerant [11, 19], or detected using so-called Razor flip-flops and corrected via safe re-execution [9]. Recent work has proposed a new timing speculation technique, referred to as TE-Drop, that significantly outperforms the two aforementioned techniques and enables up to 57% energy savings with negligible drop in classification accuracy [29]. The idea behind TE-Drop is to simply "drop out" (or skip) erroneous MAC operations instead of re-executing them, thus eliminating the performance overhead of frequent re-execution.

Empirically evaluating and comparing timing speculation methodologies for DNN accelerators is challenging (or even intractable) for several reasons. For one, gate-level timing simulations are computationally expensive and orders of magnitude slower than functional simulations. Second, DNN accelerators tend to be large: the Google Tensor Processing Unit (TPU), for example, utilizes a systolic array containing 65K ($256 \times 256$) at its core. In the Google TPU case, therefore, timing simulations have to be performed on a gate-level netlist of roughly 42 million gates. Finally, state-of-the-art DNNs perform several millions of MAC operations over their multiple layers. Putting these together, we estimate in Section 4.1 that detailed timing simulations of a state-of-the-art DNN like AlexNet on a *batch* of 256 input images takes 384 hours!

This paper presents FATE, a new methodology for fast and accurate timing error simulation of DNN accelerators. While FATE represents a general methodology, we will demonstrate FATE in the context of systolic array based DNN accelerators, which is used in the Google TPU and is one of the most popularly used architectural paradigms for DNN acceleration.

FATE builds on two separate but complementary ideas. Both ideas leverage the fact that DNN accelerators (and es-pecially systolic arrays) are large, regular arrays of identical MAC units.

1. *DNN based acceleration of timing simulations*: instead of running full gate-level timing simulations, FATE instead trains a DNN, which we refer to as DelayNet, to accurately estimate the delay of a MAC unit as a function of its inputs. DelayNet is then used along with functional simulations to evaluate the impact of timing errors on the DNN accelerator.

2. *Sampling based timing error estimation*: instead of simulating all MAC units in the DNN accelerator, FATE samples and performs timing simulations on only a subset of MAC units, for instance, a subset of columns of a systolic array, and probabilistically injects errors in the remaining MAC units at the same rate.

We show that DelayNet and sampling speed-up timing simulations for DNN accelerators by 8× and 58.57×, respectively, while introducing only 4.3%-6.17% average error in timing error estimates.

We use FATE to compare timing speculation on systolic arrays that use either 2's-complement (2C) based or sign-magnitude representation (SMR) based MAC units. We show that systolic arrays designed using SMR MACs have significantly lower timing error rates compared to their 2C counterparts and provide greater energy savings for the same classification accuracy. These results are of independent interest to DNN accelerator designers.

## 2. RELATED WORK

There is a considerable body of work on timing speculation based low-power design of digital logic [9,10,12], including recent work that has evaluated or developed new timing speculation mechanisms for DNN accelerators [14,27,29]. However, none of these works focus on frameworks to speed-up timing simulations of DNN accelerators, which is the goal of FATE.

The problem of fast timing error simulation has been addressed in prior work as well. Proposed techniques include the use of symbolic timing analysis tool [24, 25] and more recently DNN based timing estimation [13]. While Jiao et al.'s work is closest to ours, there are several important differences. First, Jiao et al. are focused on timing estimation for microprocessor modules while our target is DNN accelerators. Second, Jiao et al.'s train a classifier to predict whether given inputs result in a timing error or not. DelayNet, our proposed timing model outputs a delay value and can therefore be reused to predict timing errors across differnt clock frequency and voltage scaling values. Finally, FATE incorporates a second idea, i.e., statistical sampling of MAC operations, that is not studied in Jiao et al.'s work.

Kruijf et al. [6] propose a micro-architecture level model to estimate the overhead of timing errors, but the model still requires detailed gate-level timing simulations. We note that statistical sampling has been proposed in the past as a technique to speed-up micro-architectural simulations of multi-core processors [28], but to the best of our knowledge, we are the first to apply this methodology for timing speculative DNN accelerators.

## 3. BACKGROUND

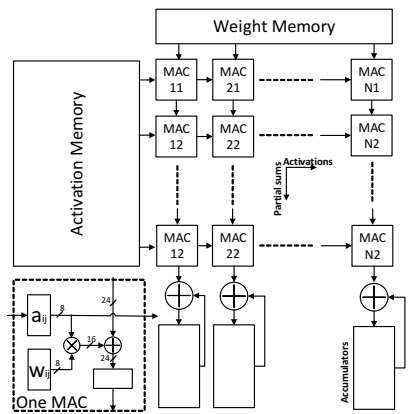

Figure 1: Systolic array based DNN accelerator based on the Google TPU [15].

This section describes the requisite background relevant to FATE. We start by briefly describing the design of DNN accelerators.

### 3.1 DNN Accelerators

A DNN is structured as a feed-forward network that contains $L$ nested layers of computation. Layer $i \in [1, L]$ has $N_i$ "neurons" whose outputs $a^i \in \mathbb{R}^{N_i \times B}$ are called activations. Here $B$ is the batch size, i.e., the number of inputs the DNN operates on simultaneously. Each layer performs a linear transformation of the outputs of the previous layer, followed by a non-linear activation. The operation of a DNN can be described mathematically as:

$$a^i = \phi\left(w^i a^{i-1} + b^i\right) \quad \forall i \in [1, L], \quad (1)$$

where $\phi : \mathbb{R}^{N_i} \to \mathbb{R}^{N_i, B}$ is each layer's activation function, $w_i \in \mathbb{R}^{N_{i-1}} \times N_i$ is the weight matrix, and $b_i \in \mathbb{R}^{N_i}$ are referred to as the bias. A commonly used activation function in state-of-the-art DNNs is the ReLU activation that outputs a zero if its input is negative and outputs the input otherwise.

Multiplying the matrix of weights $w^i$ with a matrix of activations $a^i$ is the most computationally expensive operation in DNN execution[1]. Consequently, the primary focus of DNN accelerators is in speeding-up matrix multiplications. For instance, the baseline DNN accelerator shown in Figure 1 uses a systolic array, a grid of $N \times N$ MAC units, to accelerate matrix multiplications.

The operation of a systolic array can be understood as follows. First, a matrix of $N \times N$ is loaded into the array from the weight memory, one weight per MAC unit. Next activations are read from the activation memory and flow through the array from left to right. Each MAC unit receives two inputs per clock cycle; an activation from the left and a partial sum from the north. It multiplies its incoming activation with its stored weight and adds the product to the incoming partial sum; the resulting sum is sent to the downstream MAC unit. The final MAC unit in column $i$ outputs, in successive clock cycles, the dot product of row $i$ of the weight matrix with each column in the activation matrix. The final results are stored in accumulation buffers and copied back into activation memory for subsequent computations.

---
[1]The time complexity of matrix multiplication is $O(N^2 B)$ for a DNN with $N$ neurons in each layer versus only $O(NB)$ for applying the activation function

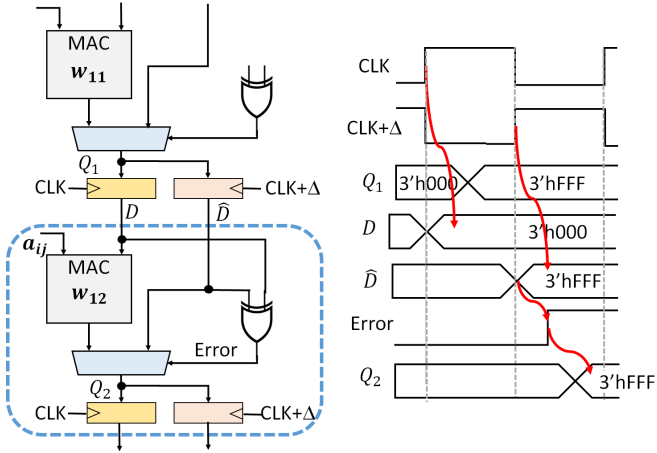

Figure 2: Block-level and timing diagram of TE-Drop proposed in [29].

### 3.2 Timing Speculation

Timing speculation was first proposed by Ernst et al. [9] as a strategy to speculatively operate digital logic at a lower voltage than that required to guarantee that timing constraints are met. While lowering voltage quadratically reduces dynamic power, it also increase the delay of combination logic. Consequently, under-volting the logic can result in timing errors. Errors are detected using double-sampling flip-flops that latch the combinational logic output at the regular clock edge and using a delayed clock. If the two outputs are different, a timing error is inferred. The input that caused the timing error is then re-executed, but this time with a slower clock so as to guarantee correct execution. We will refer to this approach as timing error detection and re-execution (TEDR).

For algorithms that are inherently error resilient, a simpler way to deal with timing errors is to simply allow them to propagate instead of incurring detection and re-execution overhead [14, 19, 26]. This is referred to as timing error propagation (TEP).

Recent work has shown that TEDR nor TEP have limited energy saving potential for DNN accelerators as they only allow the accelerators to run at relatively low timing error rates below 1% [14, 29]. While we refer the interested reader to the prior work for more details as to why this is the case, we note that Zhang et al. [29] propose a new timing speculation methodology for DNN accelerators referred to as TE-Drop.

TE-Drop detects timing errors in the same way as TEDR, but differs in the way it responds to errors. When a timing error is detected in a MAC operation, TE-Drop borrows a cycle from the subsequent MAC to correctly complete its own execution, but drops (i.e., zeroes out) the contribution of the subsequent MAC operation to the partial sum (see block diagram and timing diagram in Figure 2). Empirical results show that TE-Drop enables execution at timing error rates as high as 10% with negligible drop in classification accuracy and no performance loss. we therefore adopt TE-Drop as our baseline timing speculation mechanism in the remainder of this paper.

## 4. FATE METHODOLOGY

In this section we describe the FATE methodology. We begin by motivating the need for FATE and then describe the two new techniques that constitute FATE, i.e., DelayNet, a DNN based timing model for MAC units, and sampling based acceleration of timing simulations.

### 4.1 Motivation

To demonstrate the intractability of running detailed gate-level timing simulations of a DNN accelerator, we developed an RTL prototype of a systolic array, modeled on the Google TPU, with $256 \times 256$ MAC units and synthesized it with the OSU FreePDK 45nm Library. More details of our experimental setup can be found in Section 5. We then scheduled the four benchmark DNNs, as described in Table 1, on the prototype and ran detailed post-synthesis timing simulations using Modelsim for a batch of 256 test inputs. Table 1 shows the number of MAC operations that each DNN performs, and the simulation time required to execute a batch of test inputs in detailed timing simulation mode. Note that running AlexNet, the largest DNN we simulated, takes more than 10 days for only 256 inputs which represents only a fraction of AlexNet's test suite. This study motivates the acute need to speed-up timing error simulations for DNN based accelerators.

### 4.2 DelayNet

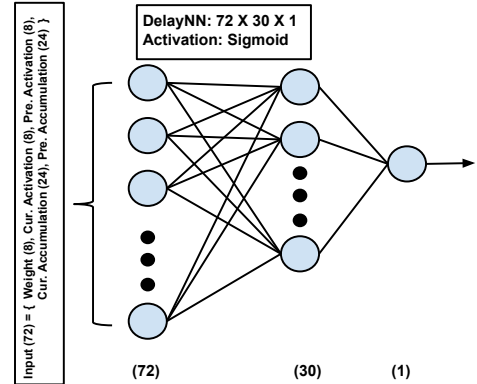

Figure 3: Architecture of DelayNet. Assuming 8-bit weights and activations and 24-bit partial sums, DelayNet takes a 72-bit input and outputs a real valued estimate of the MAC unit delay for the given inputs.

The first component of FATE is DelayNet, a DNN based timing model for MAC units. DelayNet is motivated by the observation that MAC units constitute a dominant fraction of the logic of DNN accelerators; for instance, MAC units constitute $> 99\%$ of the logic of systolic array shown in Figure 1. Furthermore, from our simulations we noted that all timing errors are triggered by MAC units since the control logic is simple and has low delay. Therefore, any method that speeds-up timing simulations for the MAC unit provides commensurate speed-ups in timing simulation of the full DNN accelerator.

Given the success of deep learning for regression tasks, we train a 2-layer fully-connected DNN with sigmoid acti-vations, referred to as DelayNet, to approximate the delay of MAC units. Ideally, the inputs to DelayNet would $m$-bit activation inputs, $n$-bit partial sum inputs and and $k$-bit weight inputs to the MAC unit from the current and previous clock periods. However, since our baseline systolic array is weight stationary [5], we do not require the weight from the previous cycle. Hence, DelayNet has a $2m + 2n + k$ bit input and a real-valued delay output. Figure 3 shows the DelayNet architecture.

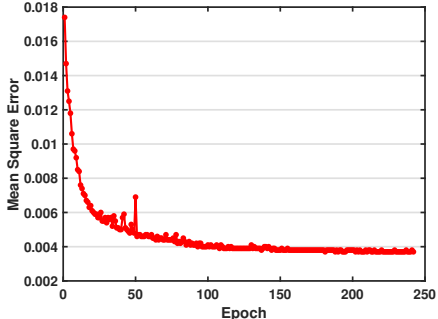

Figure 4: Training error with respect to number of training epochs. Each training epoch takes roughly 10 seconds.

DelayNet is trained using inputs and delays obtained from full timing simulations of the systolic array. During timing simulations, the systolic array is exercised with training data for the DNN benchmarks under evaluation. From the timing simulations, we extract 1M MAC input and corresponding delay pairs as training data for DelayNet. We are helped by the fact that DNN benchmarks are accompanied by large and representative training data sets that reflect the actual distribution of MAC inputs during test. Figure 4 shows evolution of training error as a function of the training epoch. We note that the error converges rapidly and the entire network is trained within 10 minutes with an root mean square error of 0.038 (we assume delay values are normalized in $[0, 1]$).

Once trained, DelayNet can be re-used across experiments. For example, in Section 5 we use DelayNet to estimate timing error probabilities and classification accuracy at multiple different voltage underscaling ratios. Furthermore, although not illustrated in this paper, DelayNet can be used to compare different DNN accelerators architectures as long as they use the same MAC design.

### 4.3 Sampling Based Acceleration

Even after speeding-up timing simulations with DelayNet, the sheer number of MAC operations performed by large DNNs can still result in prohibitive simulation time. To address this problem, we use statistical sampling [28], an approach that has already been effectively applied in speeding-up micro-architecture simulations of multi-core processors.

Specifically, we randomly sample a subset of MAC operations during DNN execution and run either full or DelayNet based timing simulations on these samples. From the samples, we then compute the average error probability over these samples $p_{err}$ (i.e., the fraction of sampled MAC operations that cause timing errors). Finally, we randomly inject timing errors in the remaining MAC units with probability $p_{err}$.

For systolic array based DNN accelerators, FATE randomly samples a subset of columns on which to perform timing simulations. Note each column in the systolic array computes the output of a different neuron. Column sampling has two advantages: (i) we expect the timing error rates of columns to be correlated since each column has the same activation inputs; and (ii) each sampled column can be simulated in parallel since there are no data dependencies between columns of the systolic array. The pseudo-code for FATE with column sampling is described in Algorithm 1.

Note that the sampling methodology is repeated per layer of DNN execution. This is because the distribution of activation inputs can change significantly from one layer to the next. Indeed, because prior work [29] has noted that DNN timing error rates vary significantly across DNN layers, further motivating the need to sample for each layer separately.

### 4.4 FATE Tool-flow

The overall FATE Tool-flow, illustrated in Figure 5, contains two phases. (i) The **training phase** maps DNN benchmarks to the synthesized systolic array netlist and performs detailed gate-level timing simulation with the Standard Delay File (SDF) in Modelsim to extract the ground truth delay for each MAC operation. With the ground truth delay, FATE trains the DelayNet as described in Section 4.2 using Tensorflow. (ii) The **prediction phase** uses inputs from the test dataset to perform the function simulations and uses sampling to extract a subset of MAC operations from which $p_{err}$ is estimated using timing simulations. Probabilistic error injection with probability $p_{err}$ is then used to determine the classification accuracy versus energy tradeoffs in the presence of timing speculation.

## 5. EMPIRICAL EVALUATION

We now evaluate the accuracy and speed-up of FATE, and demonstrate its use in enabling architectural exploration for timing speculation based DNN accelerators. We begin by describing our experimental setup.

Table 1: Simulation time for detailed gate-level timing simulations of benchmark DNNs for a test set containing 256 inputs. Also shown are the architectural parameters and total number of MAC operations for each benchmark DNN.

| Benchmarks | | 256 Input Batch | |
|---|---|---|---|
| Name | Architecture | MAC OPs | Sim. Time |
| MNIST | L1-L4 (FC): $784 \times 256 \times 256 \times 256 \times 10$ | 3.34e5 | 30 min |
| Reuters [3] | L1-L4 (FC): $2048 \times 256 \times 256 \times 256 \times 52$ | 6.69e5 | 50 min |
| TIMIT [2] | L1-L4 (FC): $1845 \times 2000 \times 2000 \times 2000 \times 183$ | 1.21e7 | 24.6 h |
| ImageNet [7] | L1-L2 (Conv): $(224, 224, 3) \times (27, 27, 64) \times (13, 13, 192)$<br>L3-L5 (Conv): $(13, 13, 384) \times (13, 13, 256) \times (6, 6, 256)$<br>L6-L8: $4096 \times 4096 \times 1000$ | 6.56e8 | 384 h |

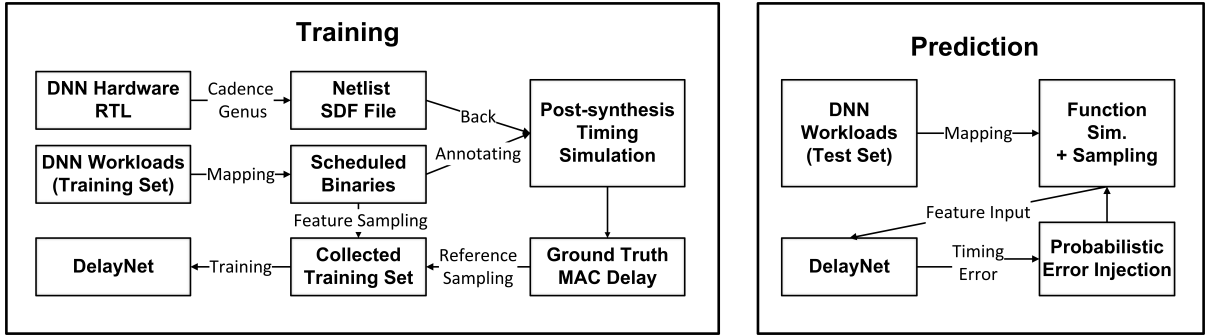

Figure 5: Diagram illustrating FATE Tool Flow.

---

**Algorithm 1: Sampling based Acceleration**

**Input** : $w^l, a^l, t_{CLK}$
**Output**: $a^{l+1}$

1 **Algorithm** Sampling($w^l, a^l, t_{CLK}$)
2    $a^{l+1} = 0$ ;
3    /*Random Selection of q Columns*/
4    $I \subset [0, N-1]$, $|I| = q$ ;
5    /*Sampling*/
6    **for** $(i=0, i \leq B, i++)$ **do**
7       **for** $j \in I$ **do**
8          **for** $(k=0, k<N, k++)$ **do**
9             **if** $MAC_{delay}(a^l_{ki}, w^l_{jk}, a^{l+1}_{ji}) > t_{CLK}$ **then**
10                $p_{err} = p_{err} + 1$;
11             **end**
12             /* Delay of MAC can be either obtained from timing simulation, or DelayNet*/
13             $a^{l+1}_{ji} += w^l_{jk} a^l_{ki}$ ;
14          **end**
15       **end**
16    **end**
17    /* Probabilistic Error Injection*/
18    **for** $(i=0, i \leq B, i++)$ **do**
19       **for** $j \notin I$ **do**
20          **for** $(k=0, k<N, k++)$ **do**
21             **if** $Bernoulli(p_{err}) == 1$ **then**
22                $a^{l+1}_{ji} = a^{l+1}_{ji}$;
23             **end**
24             **else**
25                $a^{l+1}_{ji} += w^l_{jk} a^l_{ki}$ ;
26             **end**
27          **end**
28       **end**
29    **end**
30    **return** $a^{l+1}$;
31

---

## 5.1 Setup

*DNN Benchmarks.*

We evaluate FATE on four popular DNNs: two small DNNs for MNIST digit classification and Reuters text categorization [3], and two large state-of-the-art DNNs for TIMIT speech recognition [2] and image recognition using the ImageNet dataset [7]. The parameters of each DNN benchmark is shown in Table 1. Of the four DNNs, three are multi-layer perceptrons (MLP) and the largest, AlexNet, is a convolutional neural network (CNN). Indeed, although we described FATE in the context of MLPs, it can be used to evaluate CNNs as well.

*Baseline DNN Accelerator.*

All our experiments are performed on a systolic array based DNN accelerator that closely resembles the Google TPU. The systolic array in our accelerator has 65K MAC units arranged in a square grid. Each MAC unit in the array operates on inputs represented as 2's complement (2C) signed integers with 8-bit weights and activations and 24-bit partial sums. A cycle-accurate prototype of this accelerator is implemented in fully-synthesizable Verilog and synthesized with the 45 nm OSU FreePDK technology library using Cadence Genus. Gate-level timing simulations are performed using ModelSim.

## 5.2 FATE Accuracy and Speed-up

In our experimental results, we evaluate FATE in terms of accuracy and speed-up with respect to full timing simulations which we refer to as **Full-Sim**.

- **FATE-DNN**: simulates all MAC units in the systolic array but uses DelayNet instead of full gate-level timing simulations.

- **FATE-Samp**: uses the proposed sampling methodology with sampling parameter $q = 32$. That is, 32 of the 256 columns in the systolic array are sampled and simulated fully while probabilistic error injection is used for the rest.

Note that FATE-DNN and FATE-Samp are orthogonal and can be used *together* to offer multiplicative speed-ups. Here, we characterize each technique's accuracy and speed-up separately.

Figure 6 shows the per-layer timing error rates for MNIST, Reuters and TIMIT estimated using Full-Sim, FATE-DNN and FATE-Samp. Note that we were unable to run full timing simulations for AlexNet because of the prohibitive runtime of doing so. From Figure 6 we observe that the timing error rates obtained using FATE-DNN and FATE-SIM are close to those obtained from Full-Sim. Both FATE-DNN and FATE-Samp capture two important qualitative properties of timing errors: (i) the increase in timing error rate with voltage scaling, and (ii) the fact that timing errors rates vary significantly from one layer to the next.

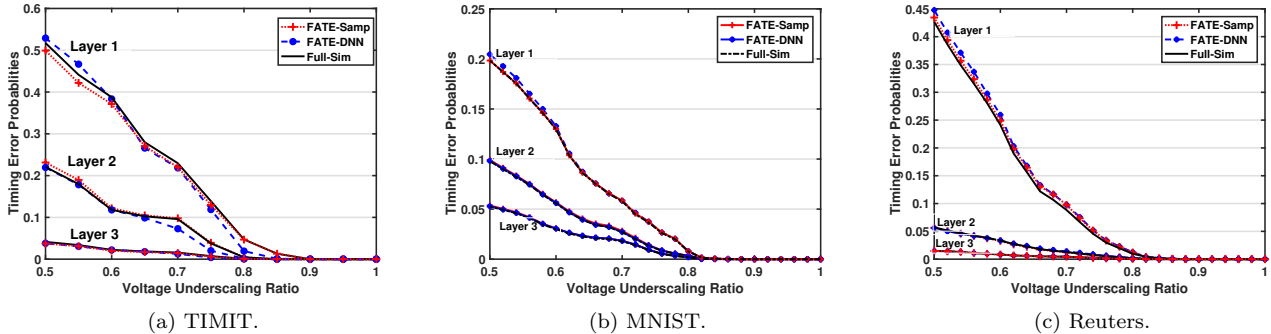

Figure 6: Timing Error Rates on TIMIT, MNIT, Reuters Test Dataset.

Table 2: FATE Accuracy and Speed-up

| Benchmark | FATE-Samp | | FATE-DNN | |
|---|---|---|---|---|
| | Error | Speedup | Error | Speedup |
| MNIST | 2.33% | 8X | 2.73% | 3X |
| Reuters | 2.16% | 8X | 4.76% | 5.8X |
| TIMIT | 4.3% | 8X | 6.17% | 58.57X |

Quantitatively, we note in Table 2 the average error in timing error estimates obtained from FATE-DNN and FATE-Samp as well as the speed-ups compared to Full-Sim. Note that that FATE-DNN speed-ups include the training time for DelayNet. We note that FATE-Samp is more accurate compared to FATE-DNN but is also slower, providing $8\times$ speed-up over Full-Sim with only 4.3% error in timing error estimates.

Next, we used FATE-DNN and FATE-Samp to evaluate the classification accuracy versus energy trade-offs enabled by the TE-Drop timing speculation scheme. Recall that as voltage is reduced, the DNN accelerator's energy consumption drops but its timing error rate increases. Consequently, a larger fraction of MAC operations are "dropped" resulting in a decrease in classification accuracy. Figure 7 plots the classification accuracy versus energy trade-off curves obtained using FATE-DNN and FATE-Samp along with the baseline results from Full-Sim for the MNIST, Reuters and TIMIT DNNs. We note that the classification accuracy estimated using FATE-DNN and FATE-Samp are always within 2% of the golden value obtained from Full-Sim.

### 5.3 Architectural Exploration Using FATE

We now illustrate how FATE enables architectural exploration for low-power, timing speculation based DNN accelerators. Specifically, we will compare two implementations of systolic array based DNN accelerator in Figure 1: the first makes use of 2's complement (**2C**) representation for weights, activations and partial sums, while the second makes use of a sign-magnitude representation (**SMR**).

The reason we believe SMR might be advantageous over 2C from a timing error perspective is as follows. Prior work and our own empirical observation show that the weights, activations and partial sums in DNNs cluster around small positive and negative values (see Figure 8c for example). In a 2C representation, small positive values are encoded using logic 1s in the least significant bits (LSBs) while small negative values are encoded using logic 1s in the most significant bits (MSBs). Hence, switching activity is distributed across the LSBs and MSBs. On the other hand, small positive and negative values are encoded identically (except the sign bit) for SMR, and hence the switching activity is focused in the LSBs.

In Figure 8a and Figure 8b we confirm this intuition using full timing simulations on TIMIT. Observe that although 2C and SMR have the same worst-case delay values[2], the mean delay of SMR for small negative weights is significantly smaller than that for 2C.

Using both full timing simulations and FATE on MNIST, Reuters and TIMIT, we confirm that SMR results in more favorable classification accuracy versus energy trade-offs compared to 2C. However, since 2C itself performs quite well on these three benchmarks, the improvements from SMR are relatively modest.

We now use FATE-Samp to compare 2C versus SMR on AlexNet, for which full timing simulations on the validation dataset are infeasible. In Figure 9(a)-(b), we plot the per-layer timing error rates for 2C and SMR implementations. In both cases Layer 1 has the highest rate of timing errors, but the timing error rates for SMR are lower than those for 2C. The timing error rates for other layers are comparable.

Figure 9(c) compares the classification accuracy vs. energy trade-offs obtained using TE-Drop based timing speculation on 2C and SMR implementations[3]. We observe that SMR significantly outperforms 2C, resulting in up to 14.45% higher classification accuracy for the same energy savings. Conversely, SMR enables greater energy savings for similar classification accuracy. These analyses are enabled by the simulation speed-ups that FATE provides.

### 6. CONCLUSION AND FUTURE WORK

In this paper we have presented FATE, a methodology for fast and accurate timing error rate estimation of time speculation based low-power DNN accelerators. Using two novel techniques, i.e., DNN based delay prediction and statistical sampling of MAC operations, FATE is able to reduce the run-time of timing simulations for AlexNet from 384 hours to between 7 to 48 hours, depending on the approach used. At the same time, FATE's timing error rates and DNN classification accuracy estimates are within 6% and 2% of those obtained from full simulations. We have used FATE to compare 2C and SMR based implementations of a DNN accel-

---
[2] We synthesized both with the same timing target for a fair comparison.
[3] As is common practice, we report both the Top 5 and Top 1 accuracy for ImageNet

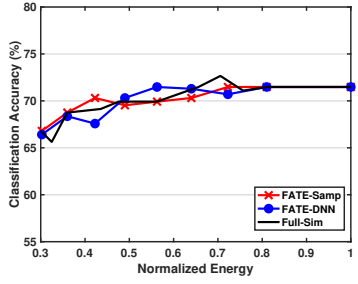
(a) TIMIT Accuracy vs. Energy validation.

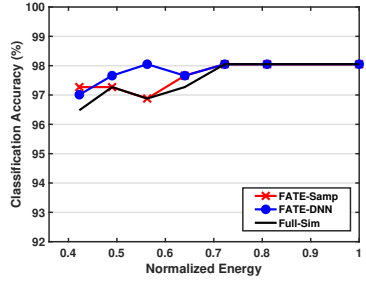
(b) MNIST Accuracy vs. Energy validation.

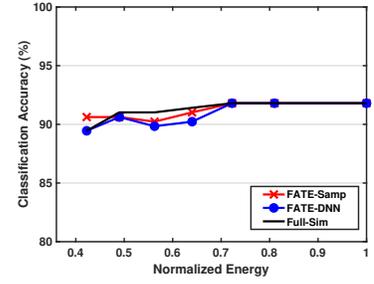
(c) Reuters Accuracy vs. Energy validation.

Figure 7: Accuracy Energy Tradeoff Validation (a) TIMIT, (b) MNIST, and (c) Reuters.

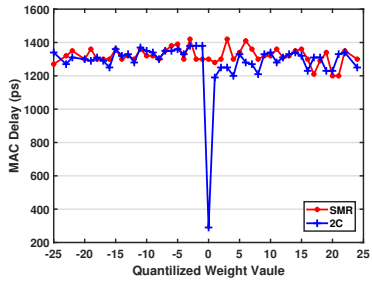
(a) Maximum MAC Delay versus Weight.

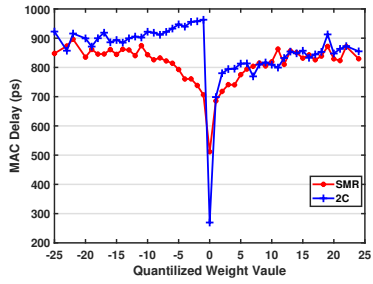
(b) Mean MAC Delay versus Weight.

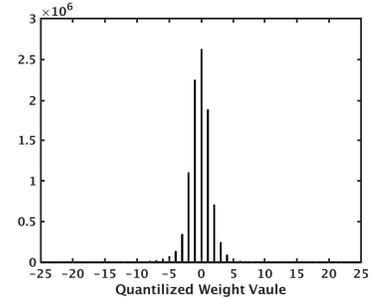
(c) Weight Distribution.

Figure 8: 2C Vs SMR MAC Design, TIMIT Dataset, (a) Max MAC Delay per Weight, (b) Mean MAC Delay per Weight, and (c) Weight Distribution.

erator modeled on the Google TPU and show that the SMR implementation provides more favorable accuracy vs. energy trade-offs compared to the 2C implementation.

As future work, we would like to provide a rigorous statistical analysis of FATE, and leverage FATE to compare a broader range of DNN accelerator architectures under timing speculation.

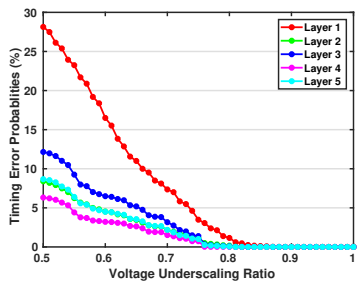
(a) 2s comp Timing Errors of AlexNet.

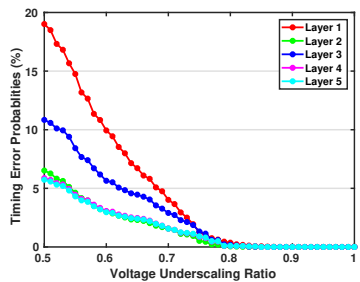
(b) SMR Timing Errors of AlexNet.

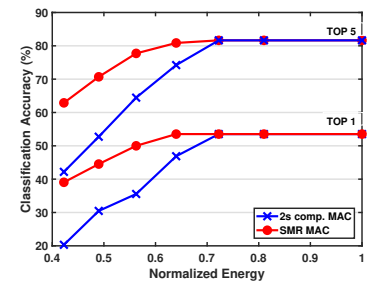
(c) Energy Accuracy tradeoff AlexNet.

Figure 9: ImageNet Dataset (a) 2's compliment MAC, (b) SMR MAC, and (c) Energy Accuracy tradeoff.